\documentclass[conference]{IEEEtran}
\IEEEoverridecommandlockouts
\usepackage{cite}
\usepackage{amsmath,amssymb,amsfonts}
\usepackage{graphicx}
\usepackage{textcomp}
\usepackage{xcolor}
\usepackage{mathtools}
\usepackage{colortbl}
\usepackage{notoccite}
\usepackage{booktabs} 
\usepackage{hyperref}

\def\BibTeX{{\rm B\kern-.05em{\sc i\kern-.025em b}\kern-.08em
    T\kern-.1667em\lower.7ex\hbox{E}\kern-.125emX}}

\newcommand{\be}{\begin{eqnarray}}
\newcommand{\ee}{\end{eqnarray}}

\newcommand\fnurl[2]{%
  \href{#2}{#1}\footnote{\url{#2}}%
}

\begin{document}

\title{A Flexible Framework for Anomaly Detection via Dimensionality Reduction
}

\author{\IEEEauthorblockN{Alireza Vafaei Sadr}
\IEEEauthorblockA{\textit{School of Physics,} \\
\textit{Institute for Research in Fundamental Sciences (IPM),}\\
\textit{P. O. Box 19395-5531, Tehran, Iran}\\
\textit{D\'epartement de Physique Th\'eorique and} \\
\textit{Center for Astroparticle Physics,} \\
\textit{University of Geneva}\\
vafaei@ipm.ir}
\and
\IEEEauthorblockN{Bruce A. Bassett}
\IEEEauthorblockA{\textit{African Institute for Mathematical Sciences, Cape Town}\\
\textit{South African Radio Astronomy Observatory}\\
\textit{South African Astronomical Observatory} \\
\textit{Mathematics Department, University of Cape Town} \\
bruce.a.bassett@gmail.com}
\and
\IEEEauthorblockN{M. Kunz}
\IEEEauthorblockA{\textit{D\'epartement de Physique Th\'eorique and Center for Astroparticle Physics} \\
\textit{University of Geneva}\\
Geneva, Switzerland \\
martin.kunz@unige.ch}
}

\maketitle

\begin{abstract}
Anomaly detection is challenging, especially for large datasets in high dimensions. Here we explore a general anomaly detection framework based on dimensionality reduction and unsupervised clustering. We release  
DRAMA, a general python package that implements the general framework with a wide range of built-in options. We test DRAMA on a wide variety of simulated and real datasets, in up to 3000 dimensions, and find it robust and highly competitive with commonly-used anomaly detection algorithms, especially in high dimensions. The flexibility of the DRAMA framework allows for significant optimization once some examples of anomalies are available, making it ideal for online anomaly detection, active learning and highly unbalanced datasets. 
\end{abstract}


\section{Introduction}

Anomaly and Novelty Detection is an important area of machine learning research and critical across a spectrum of applications which stretch from humble data cleaning to the discovery of new species or classes of objects. An example of the latter application is provided in astronomy by the LSST\footnote{https://www.lsst.org/} and SKA\footnote{https://www.skatelescope.org/}, the next-generation optical and radio telescopes which are so much more powerful than existing facilities that they are expected to observe completely new types of celestial objects lurking in the torrent of data in the 100PB-10EB range. Other real-world applications include adverse reaction identification in medicine, fraud detection, terrorism, network attacks and abnormal customer behaviour 
\cite{williams1997mining,dumouchel1998fast,pham2016anomaly}. 


Despite the wide range of approaches, the problem is still one of the most challenging areas of machine learning. The No Free Lunch (NFL) theorems\footnote{http://no-free-lunch.org/} imply that no ``best" anomaly detection algorithm exists across all possible anomalies, classes, data and problems. For any algorithm it is possible to construct anomaly attacks that deceive the algorithm by exploiting the features learned in the process of training the algorithm. 


One might be tempted to try to circumvent this aspect of the NFL theorems by building a very large number of features in the hope that some features will, by chance, be sensitive to the anomalous signal. Unfortunately, significantly increasing the number of features leads to the curse of dimensionality \cite{curse}: the performance of most machine learning algorithms deteriorate as the dimensionality of the feature spaces increases dramatically. The key reasons for the ``curse" are that distance measures  become less and less informative  \cite{beyer1999nearest} and feature space volume grows exponentially in higher dimensions. 

Contrary to the dual challenges posed by the NFL theorems and the curse of dimensionality, humans are relatively good at anomaly detection in the real world and have the ability to learn from a single example. It is therefore reasonable to believe that there exist optimal anomaly detectors for subclasses of anomalies relevant to the real world. Most physically-relevant functions are fairly smooth and can be efficiently compressed \cite{lin2017does}.  This inspires our search for ``better'' algorithms and is the key context of the present work. Application to the case of relatively smooth functions and real-world anomaly datasets is how we judge our anomaly detection algorithm, which we call the Dimensionality Reduction Anomaly Meta-Algorithm (DRAMA). DRAMA\footnote{DRAMA is based on the popular scikit-learn \cite{scikit-learn} and TensorFlow \cite{tensorflow} packages and comes with a Jupyter notebook interface for ease of use.} is released as a python package \footnote{https://github.com/vafaei-ar/drama}.


Comparison of our algorithm, DRAMA, with a large number of existing algorithms is computationally infeasible. We therefore pick two popular general algorithms to benchmark DRAMA against:  Local Outlier Factor (LOF) \cite{breunig2000lof} and Isolation Forest (iForest) \cite{liu2008isolation}.  Benchmarks are performed both against simulated data and a collection of real-world anomaly datasets. 
The outline of this paper is as follows: in section \ref{sec:drama} we outline DRAMA while section \ref{dataset} describes the simulated and experimental datasets and metrics. Results and discussion are presented in section \ref{results}. 


\section{ The DRAMA Algorithm}\label{sec:drama}

\begin{figure}
\begin{center}
\includegraphics[scale=0.28]{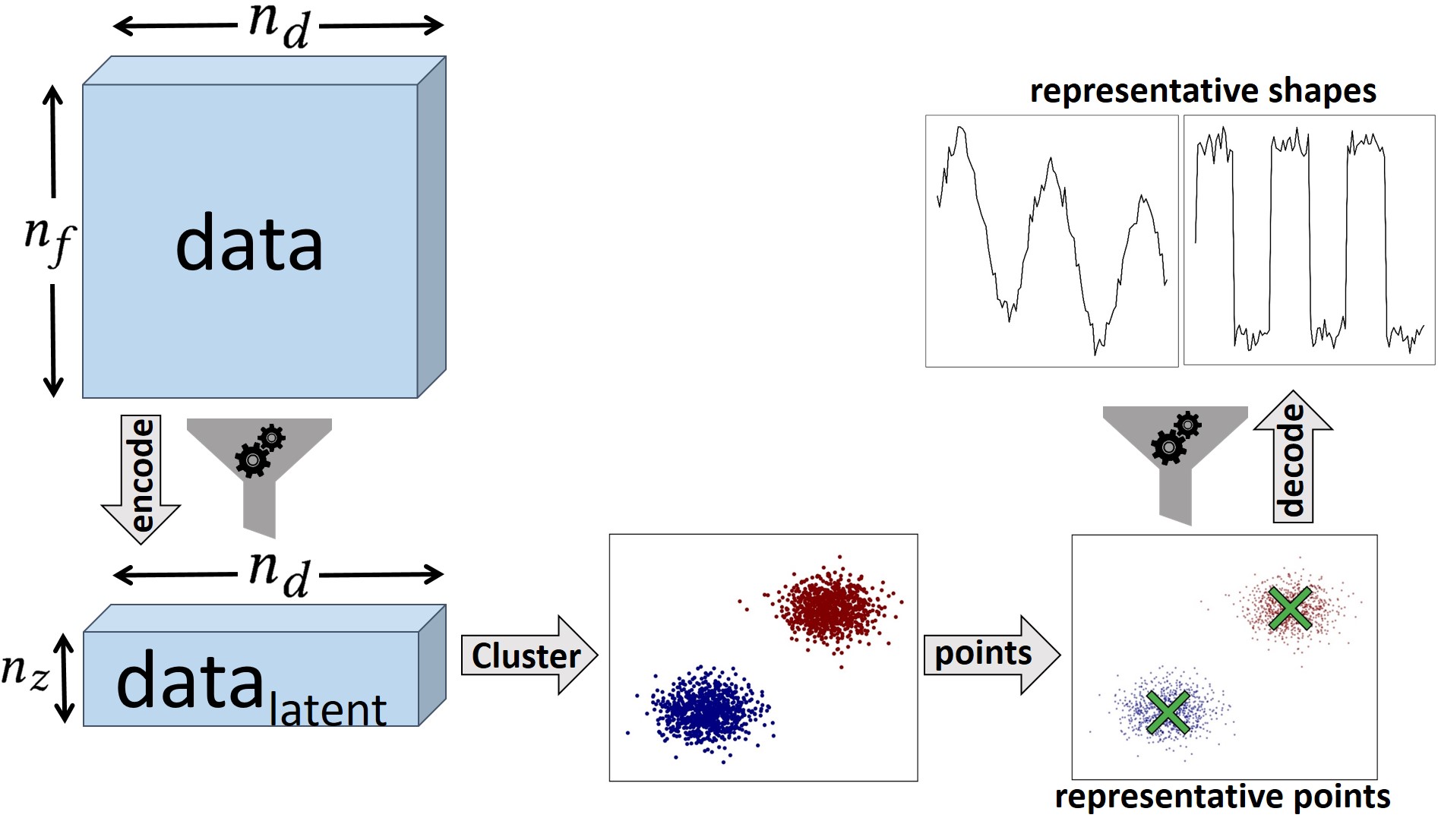}
\end{center}
\caption{Schematic of the DRAMA framework. First dimensionality reduction (left) is performed on the $n_d$ data points down from $n_f$ to $n_z$ features in the latent space. Unsupervised clustering then splits the data into clusters (here two). The covariance matrix of each label is calculated in the original feature space for the Mahalanobis distance and a prototype is extracted for cluster. This prototype (main component) is then decoded back into the original space (top right) and comparison between the prototypes and test data is performed. Finally anomalies are ranked by their maximum distance to the closest prototype in each case.}
\label{pipeline}
\end{figure}

Our algorithm -- Dimensional Reduction Anomaly Meta Algorithm (DRAMA) -- consists of four main steps: 
(i) dimensionality reduction ({\em encoding}) of data to a lower-dimensional space, followed by 
(ii) clustering to find the main prototypes in the data, 
(iii) uplifting to the original space ({\em decoding}; optional) and finally  
(iv) distance measurements between the test data and the prototypes (the main clustered components) to rank potential anomalies. These steps are illustrated in Fig.\ref{pipeline} and are discussed in turn in the following. 

\subsection{Dimensionality reduction}\label{sec:drm}

This is the first step of DRAMA procedure. Assume $\textbf{X}$ as a $n_f$-dimensional feature vector. Dimensionality reduction translates $\textbf{X}$ into another vector $\textbf{z}$ in an $m$-dimensional space, where $m \ll n$. Good reductions keep as much important information as possible while removing noise and irrelevant information, efficiently {\em encoding} the data. In general the reduction will result in loss of information, but is very useful when one wants to work with or visualize the data in lower dimensions or attempts to combat the curse of dimensionality. The inverse process of lifting back up to the original space, i.e. going from $\textbf{z}$ to  $\textbf{X}$ will be referred to as {\em decoding} hereafter. DRAMA uses both encoding and decoding, with primary component extraction performed in between.  

The current version of DRAMA comes with 5 builtin Dimensionality Reduction Techniques (DRT)\footnote{DRAMA is modular and easy to extend to any other DRT.}:

\begin{itemize}
\item{Independent Component Analysis (ICA) \cite{hyvarinen2004independent,naik2012introduction}}
\item{Non-negative Matrix factorization (NMF) \cite{berry2007algorithms}}
\item{Autoencoders (AE) \cite{hinton1994autoencoders}\footnote{All the results of this paper using neural networks are produced using fully connected architectures where the number of units in each layer are chosen to be $(n_f,n_f/2,2,n_f/2,n_f)$ respectively where $n_f$ is the number of features. The learning was $0.001$ and RELU was the chosen activation function. More detailed information is available in the released package.}}
\item{Variational AutoEncoders (VAE) \cite{kingma2013auto,rezende2014stochastic}}
\item{Principal Component Analysis (PCA) \cite{pearson1901liii,hotelling1933analysis}}
\end{itemize}

\subsection{Prototype extraction}\label{extraction}

For the second and third steps in DRAMA we now explain prototypes and how they can be extracted. Having encoded the features down to a low-dimensional latent space (here we used 2D) using a choice of DRT, we can efficiently perform unsupervised clustering to detect clusters. 
Then the primary shapes -- the prototypes or archetypes -- in the data can be captured by the centers of the detected clusters (see e.g. \cite{biehl2016prototype}). We could perform the clustering directly in the high-dimensional feature space before encoding, but this is often computationally unfeasible. Prototype detection can use any of the many existing clustering algorithms. In this study we report results using agglomerative clustering \cite{day1984efficient} because experiments showed that it was superior to other methods like K-means \cite{hartigan1979algorithm} 
\footnote{Several other clustering algorithms are included in the DRAMA package however, allowing the users to choose for themselves.}.

\begin{figure}
\begin{center}
\includegraphics[scale=0.27]{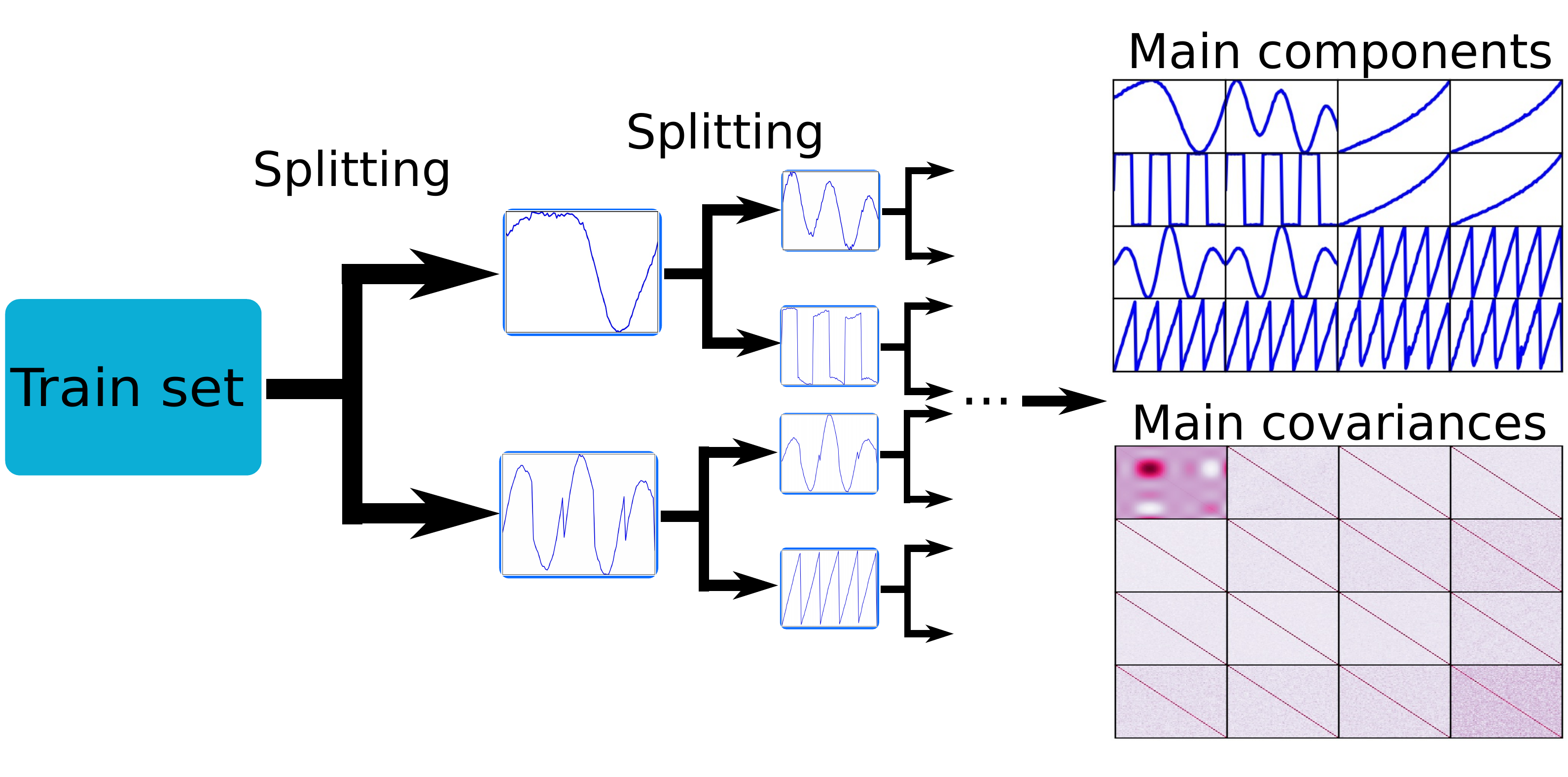}
\end{center}
\caption{Illustrating prototype extraction process in DRAMA. Splitting can be done iteratively to extract more detailed information about the different shapes in the data, allowing more prototypes to be extracted if the inlier data has a large amount of variation.}
\label{split}
\end{figure}

As illustrated in Fig.\ref{split}, the data is hierarchically split into 
$2^{n_s}$ detected prototypes.  We are not concerned with  whether the $2^{n_s}$ clusters represent true subclasses; $n_s$ is simply a hyperparameter designed to find anomalies. 
Having found the $2^{n_s}$ clusters we select the centre\footnote{Here taken to be the mean of each cluster.} of each cluster and can now choose whether or not to decode it to the original, high-dimensional, feature space. Empirically we find that decoding to the original space gives better results. Either way we now have $2^{n_s}$ prototypical components representative of the ``average” (inlier) data. 

\subsection{Identifying Anomalies}\label{sec:outlier}


Having decoded (or encoded) the prototypes, the next step is computing the distance between them and each test data point. This requires a metric, $d(x_i, c_j )$, where $x_i$ are test data points and $c_j$ are the prototypes. For any choice of metric the predicted anomalies are then ranked by  $\rm{max}_i \{ \rm{min}_j \{ d(x_i, c_j ) \} \}$.   The choice of metric is another flexibility of DRAMA. Currently DRAMA includes nine different metrics,
given in Table (\ref{metrictable}). 

\begin{table}
\centering
\caption{Metric options available in DRAMA: $d(u,v)$ is the distance between data points $u$ and $v$; $\bar{u}$ is the average of $u$. }
\label{metrictable}
\begin{tabular}{ll}
\toprule
Metric & definition\\
\midrule
L1 &  $\sum_i {\left| u_i - v_i \right|}$  \\
\midrule
L2 &  ${||u-v||}_2$  \\
\midrule
L4 &  ${||u-v||}_4$   \\
\midrule
wL2 &  ${||\frac{u-v}{\sigma}||}_2$  \\
\midrule
wL4 &  ${||\frac{u-v}{\sigma}||}_4$   \\
\midrule
Bray-Curtis &  $\sum{|u_i-v_i|} / \sum{|u_i+v_i|}$\\
\midrule
Chebyshev & $\max_i {|u_i-v_i|}$ \\
\midrule
Canberra & $\sum_i \frac{|u_i-v_i|} {|u_i|+|v_i|}$\\
\midrule
correlation &  $1 - \frac{(u - \bar{u}) \cdot (v - \bar{v})} {{||(u - \bar{u})||}_2 {||(v - \bar{v})||}_2}$  \\
\midrule
Mahalanobis &  $\sqrt{ (u-v) C^{-1} (u-v)^T }$  \\
\bottomrule
\end{tabular}
\end{table}



\section{Datasets for Testing DRAMA}\label{dataset}

Because of the NFL theorems it is always possible to construct anomaly datasets that make any algorithm look better than any other algorithm. To avoid this we first designed two simulated anomaly detection challenges and then blind-selected several real-world anomaly benchmarks to evaluate DRAMA on, all of which we now describe. 

\subsection{Simulated challenges}

\begin{figure}
\begin{center}
\includegraphics[scale=0.37]{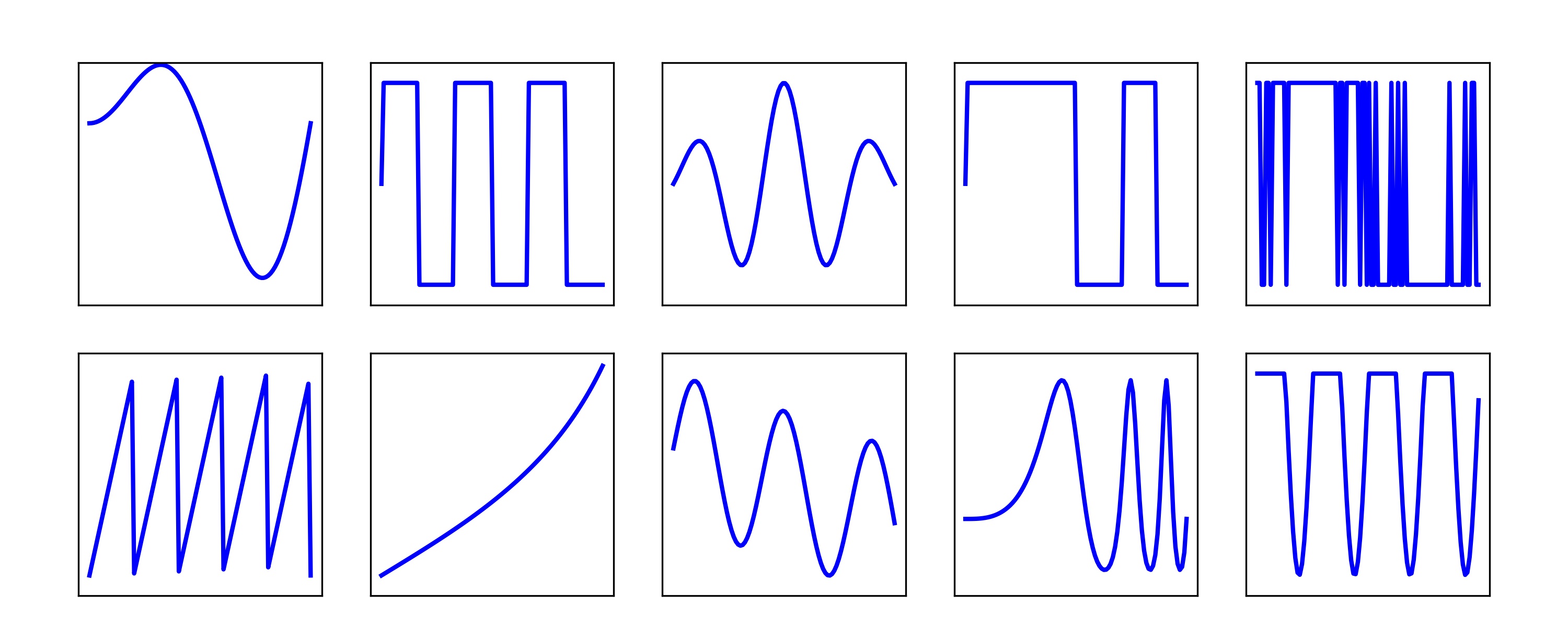}
\end{center}
\caption{The base classes used for the simulated time-series challenges. Noise and a variety of anomalies were then added on top of these base classes.}
\label{data_org}
\end{figure}

There are an infinite number of ways to simulate inliers and anomalies. In this work we consider $10$ different classes of continuous ``time series", shown in Fig.\ref{data_org}, representing a wide range of behaviours one might find in the real world.

These $10$ shapes were perturbed by random Gaussian noise and by random scaling in both the $x$ and $y$ directions. Then we designed two anomaly detection challenges, each with two sub-challenges which differ only in the dimensionality of the data. The challenge details are explained as follows: 

\begin{itemize}

\item{{\bf Challenge-I: Compact Anomalies}}

In the first challenge compact Gaussian ``bump" anomalies were added to the $10$ classes with random location, and amplitude chosen in the range $0.3-0.4$ and width in the range $0.08-0.1$. There were $1000$ inliers and $50$ anomalies for each chosen shape. Finally the noise characteristics were drawn from a mean zero Gaussian with $\sigma=0.3$, comparable to the anomaly amplitude.  This task is broken into two sub-challenges, labeled {\bf C-Ia} and {\bf C-Ib}, which differ only in the dimensionality of the data. For  {\bf C-Ia (C-Ib)} we chose $n_f = 100 (3000)$ respectively.



\item{{\bf Challenge-II}}

The second challenge uses $9$ of the shapes in Fig. (\ref{data_org}) to produce 500 inliers with the remaining shape used to produce 50 anomalies. We permute the choice of class used for the anomalies to enhance robustness. Uncorrelated Gaussian noise ($\mu=0$, $\sigma= 0.8$) is added in all cases. As before, this challenge is split into two sub-challenges, labeled {\bf C-IIa} and {\bf C-IIb}, which again differ only in the dimensionality of the data:  $n_f = 100, 3000$ respectively.





\end{itemize}



\subsection{Real datasets}


In this study we also used $20$ real-world datasets \footnote{We were limited to $20$ by computational resources.} chosen at random from the \fnurl{ODSS}{http://odds.cs.stonybrook.edu/} database. This is a standard testbed for outlier detection and has been used in many earlier works, including \cite{aggarwal2015theoretical,sathe2016lodes,rayana2016less}. A summary of the different datasets used, including the dimensionality of the data and number of inliers and outliers, is shown in Table (\ref{data_summary}).

\begin{table}
\centering
\caption{Real-world dataset summary from the ODSS Benchmark.}
\label{data_summary}
\setlength\tabcolsep{1.5pt}
\begin{tabular}{l | l | l | l}
\toprule
    Dataset & \# points & \# dim. & \# outliers \\
\midrule
      lympho &       148 &      18 &           6 \\
     breastw &       683 &       9 &         239 \\
        wine &       129 &      13 &          10 \\
   vertebral &       240 &       6 &          30 \\
       glass &       214 &       9 &           9 \\
        pima &       768 &       8 &         268 \\
     thyroid &      3772 &       6 &          93 \\
  ionosphere &       351 &      33 &         126 \\
      cardio &      1831 &      21 &         176 \\
         wbc &       378 &      30 &          21 \\
  arrhythmia &       452 &     274 &          66 \\
      vowels &      1456 &      12 &          50 \\
   satellite &      6435 &      36 &        2036 \\
  satimage-2 &      5803 &      36 &          71 \\
   optdigits &      5216 &      64 &         150 \\
 mammography &     11183 &       6 &         260 \\
     shuttle &     49097 &       9 &        3511 \\
       mnist &      7603 &     100 &         700 \\
   pendigits &      6870 &      16 &         156 \\
        musk &      3062 &     166 &          97 \\
\bottomrule
\end{tabular}
\end{table}



\subsection{Scoring metrics}

To test for robustness all the algorithms were run 10 times on each test dataset. We then report the means and best performances for two relevant metrics suited for anomaly detection, namely: area under the ROC Curve (AUC)
and Rank-Weighted Score (RWS) \cite{roberts2019bayesian}. Given a ranked list of length $N$ of the most likely outliers, the RWS is defined by:
\be
\mbox{RWS} = \frac{1}{N(N+1)}\sum_{i=1}^N w_i I_i
\ee
where the weight $w_i \equiv N + 1 - i$ and is large if $i$ is small and decreases to unity for $i = N$. Here  $I_i$ is an indicator function which is unity if the i-th object is an outlier and $0$ otherwise and the sum is over the top $N$ anomaly candidates. Here we choose $N$ to be the number of anomalies. The RWS rewards algorithms whose anomaly scores correlate well with the true probability of being an anomaly.  


\section{Results and Discussion}\label{results}

Since DRAMA is, by design, very flexible, it is actually a large number of related algorithms, differing by choice of DRT, clustering, metric etc... As a result, a DRAMA algorithm beat LOF and i-Forest on every simulated data challenge and on 17 out of 20 real-world challenges in terms of AUC. For the problems we study, the cityblock metric and AE \& NMF DRTs are the most successful on average. Because of the NFL theorems, DRAMA's superiority cannot hold in general of course, but the results show that if one DRT or metric does not perform well, another one likely will.  

\begin{figure}
\begin{center}
\includegraphics[scale=0.265]{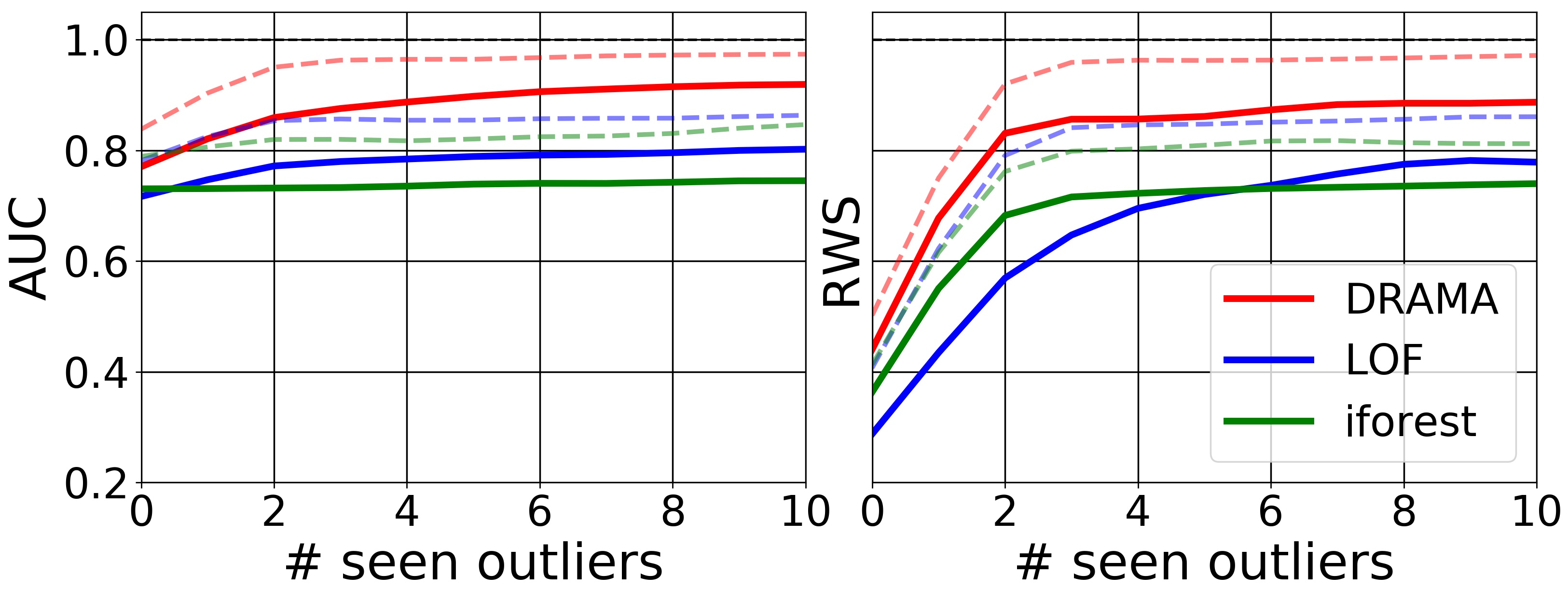}
\includegraphics[scale=0.265]{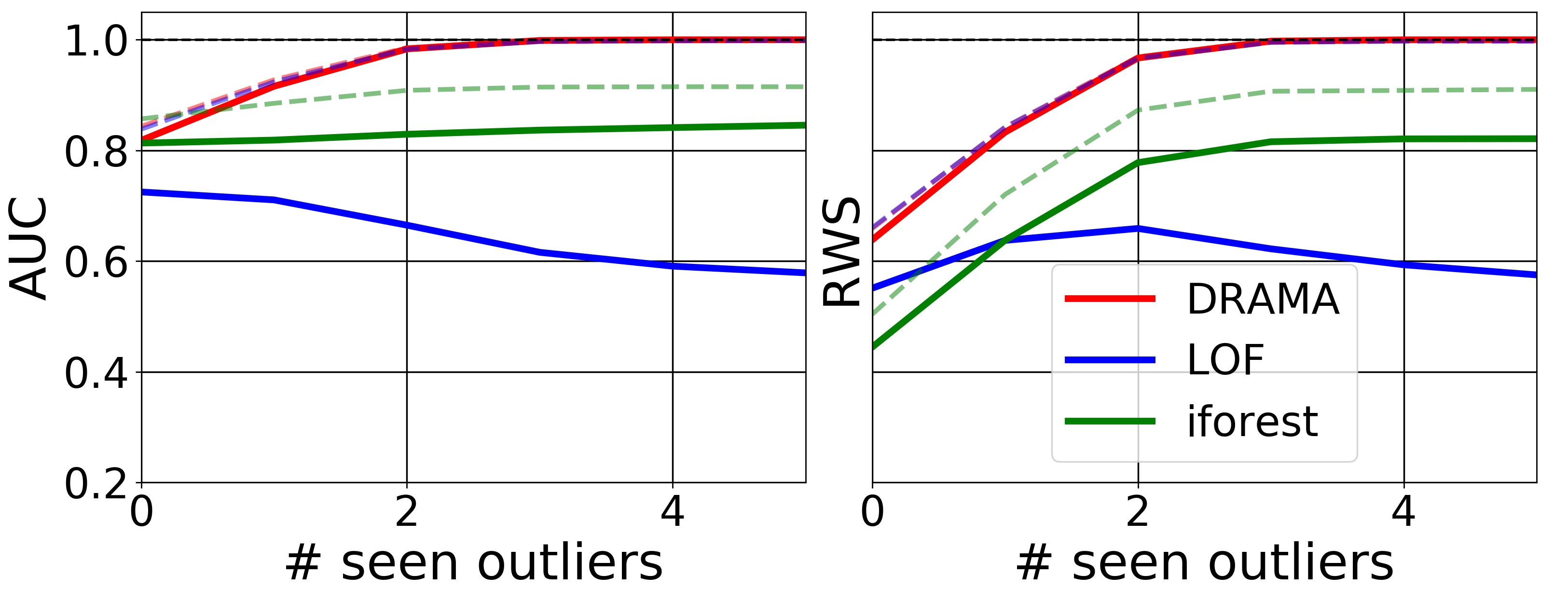}
\end{center}
\caption{In the compact anomaly challenges, ({\bf C-Ia, C-Ib}) DRAMA is far superior to both other algorithms for $n_f=100$ (top) and $n_f=3000$ (bottom), for all numbers of seen anomalies. The solid line is average performance while the dashed line is the maximum performance.}
\label{fig:ev_res}
\end{figure}

\begin{figure}
\begin{center}
\includegraphics[scale=0.265]{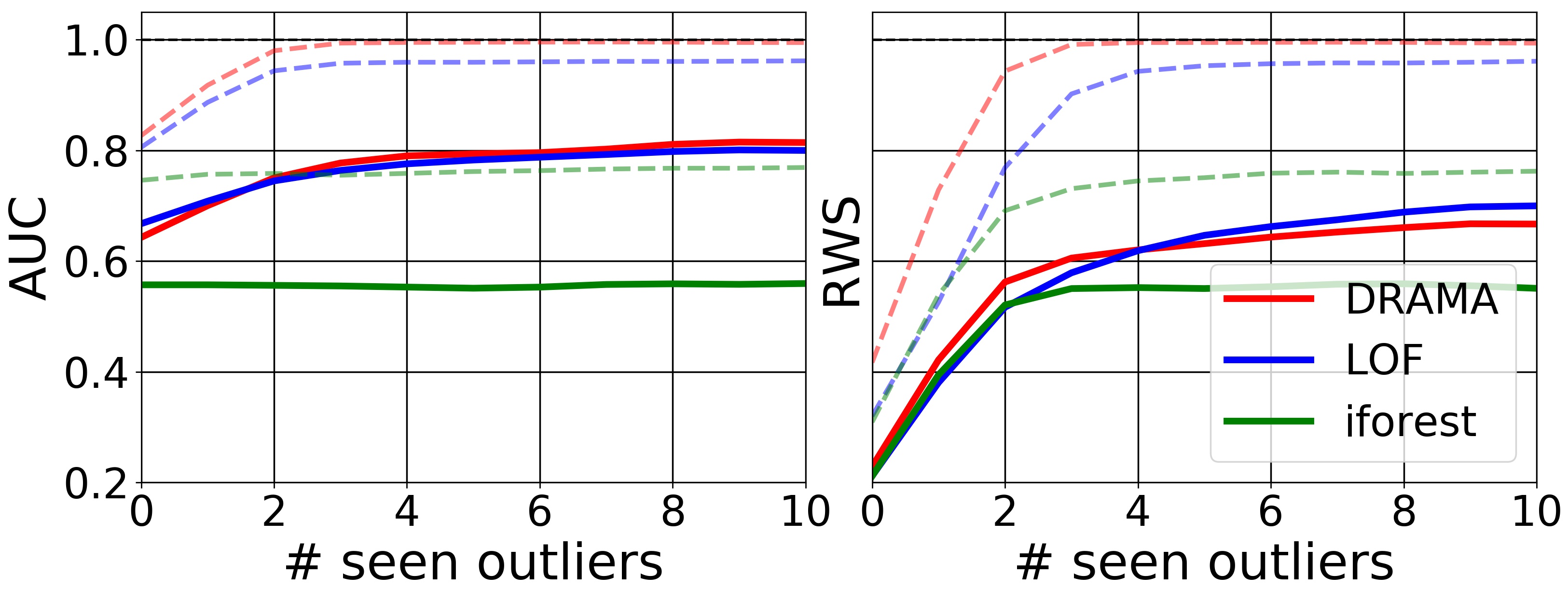}
\includegraphics[scale=0.265]{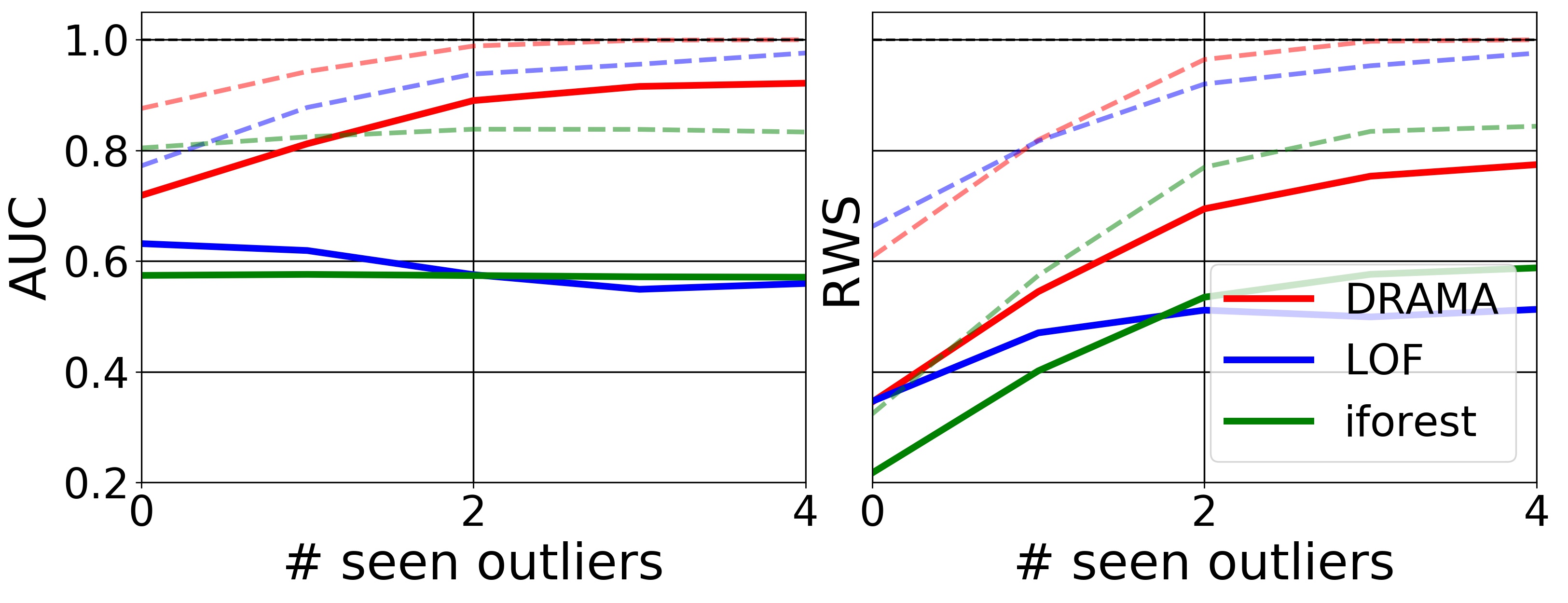}
\end{center}
\caption{In the 2nd challenge,  DRAMA is again equal to or superior to both LOF and iForest for $n_f=100$ (top, {\bf C-IIa}) and  $n_f=3000$ (bottom, {\bf C-IIb}). DRAMA particularly shines in the high-dimensional case. 
The solid line is average performance and the dashed line is the maximum performance, where DRAMA outperforms the other algorithms.}
\label{fig:mix_res}
\end{figure}

\begin{figure}
\begin{center}
\includegraphics[scale=0.265]{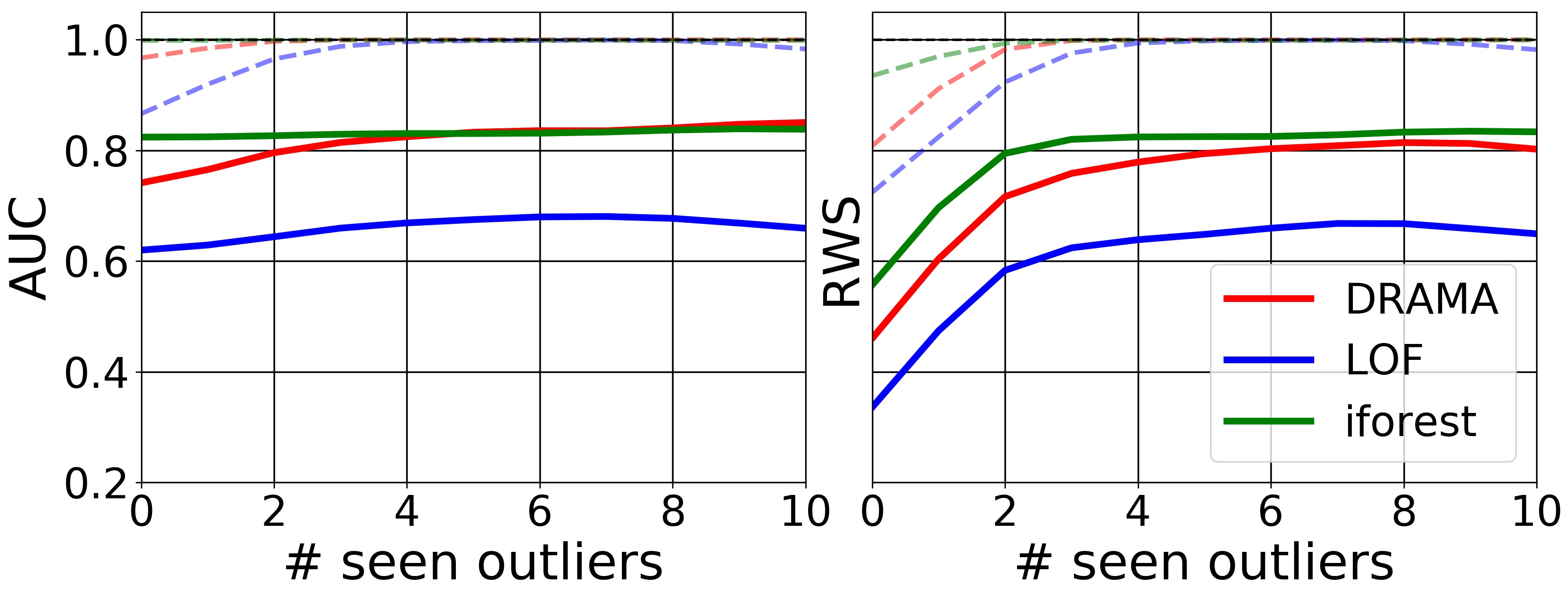}
\end{center}
\caption{Average (solid lines) and best (dashed lines) performance on the 20 real datasets where iForest and DRAMA outperform LOF. All the datasets have dimensionality less than $300$. }
\label{fig:event}
\end{figure}

The flexibility of the DRAMA framework is particularly useful when one has  seen a few anomalies or outliers. In this case one can learn the best DRT-clustering-metric  combination to allow optimal detection of the anomalies. To illustrate this capability we give DRAMA, LOF and i-Forest the ability to learn from a variable number of seen anomalies/outliers. This is used to select optimal hyperparameters for all the algorithms. The results for the two simulated challenges are shown in Fig. \ref{fig:ev_res} and Fig. \ref{fig:mix_res} and for the real datasets in Fig. \ref{fig:event}. The figures show both the mean and best results over $10$ runs for each of the challenges and for each algorithm. 

While LOF and i-Forest are competitive on the simulated challenges in low dimensions ($n_f = 100$) DRAMA particularly shines in high dimensions ($n_f = 3000$). On the real-world datasets we considered which have small numbers of points and dimensions $< 300$ the performance of DRAMA and i-Forest are comparable and significantly better than LOF.

\section{Conclusions}\label{conclusion}
Anomaly detection is a challenge particularly in the context of and high-dimensional datasets. Here we describe DRAMA, a general python package that uses a range of linear and nonlinear dimensionality reduction transformations, followed by unsupervised clustering to identify prototypes. Potential anomalies are then identified by their distance to the learned prototypes. Currently DRAMA includes five dimensionality reduction algorithms including neural network methods (AE and VAE) along with more standard methods (PCA, ICA, NMF). DRAMA also comes with nine options for the metric. 

We evaluated DRAMA performance against the commonly-used algorithms Isolation Forest (i-Forest) and Local Outlier Factor (LOF), averaging different hyperparameter configurations. The large flexibility inherent in DRAMA is particularly attractive in the case of supervised/online anomaly detection where there are known examples of the anomaly of interest because one can optimise for the best DRT-clustering-metric and hyperparameter combination to detect anomalies. 

We evaluated DRAMA on a wide variety of simulated and real datasets of up to 3000 dimensions. DRAMA particularly excelled on the simulated time-series data, winning every challenge.  On the very inhomogeneous and fairly low-dimensional real-world datasets we tested, DRAMA was highly competitive with LOF and i-Forest, showing that dimensionality reduction and clustering is a valuable approach to anomaly detection. 

Finally we note that it would be interesting to optimize DRAMA for novelty detection and anomaly detection in images. 

\section*{Acknowledgment}

We thank Emmanuel Dufourq, Ethan Roberts and Golshan Ejlali for discussions and particularly Michelle Lochner for insightful discussions and detailed comments on the draft. AVS acknowledges the hospitality and funding of SARAO and AIMS.  The numerical computations were carried out on Baobab at the computing cluster of the University of Geneva.

\bibliographystyle{IEEEtran}
\bibliography{bib}

\end{document}